1

# Two-view 3D Reconstruction for Food Volume Estimation

Joachim Dehais, *Student Member, IEEE*, Marios Anthimopoulos, *Member, IEEE*, Sergey Shevchik, Stavroula Mougiakakou, *Member, IEEE*

*Abstract*— The increasing prevalence of diet-related chronic diseases coupled with the ineffectiveness of traditional diet management methods have resulted in a need for novel tools to accurately and automatically assess meals. Recently, computer vision based systems that use meal images to assess their content have been proposed. Food portion estimation is the most difficult task for individuals assessing their meals and it is also the least studied area. The present paper proposes a three-stage system to calculate portion sizes using two images of a dish acquired by mobile devices. The first stage consists in understanding the configuration of the different views, after which a dense 3D model is built from the two images; finally, this 3D model serves to extract the volume of the different items. The system was extensively tested on 77 real dishes of known volume, and achieved an average error of less than 10% in 5.5 seconds per dish. The proposed pipeline is computationally tractable and requires no user input, making it a viable option for fully automated dietary assessment.

## I. Introduction

The worldwide prevalence of diet-related chronic diseases such as obesity (12%) and diabetes mellitus (9.3%) has reached epidemic proportions over the past decades [1], [2], resulting in about 2.8 and 5 million deaths annually and numerous co-morbidities. This situation raises the urgent need to develop novel tools and services for continuous, personalized dietary support for both the general population and those with special nutritional needs.

Traditionally, patients are advised to recall and assess their meals for self-maintained dietary records and food frequency questionnaires. Although these methods have been widely used, their accuracy is questionable, particularly for children and adolescents, who often lack motivation and the required skills [3], [4]. The main reported source of inaccuracies is the error in the estimation of food portion sizes [5], which averages above 20% [4][6][7]. Even well-trained diabetic patients on intensive insulin therapy have difficulties [8].

Smart mobile devices and high-speed cellular networks have permitted the development of mobile applications to help users assess food intake. These include electronic food diaries, barcode scanners to retrieve the nutritional content, image logs [9], and remotely available dieticians [10]. Concurrently, the recent advances in computer vision have enabled the development of automatic systems for meal image analysis [11]. In a typical scenario, the user takes one or more images or even a video of their meal, and the system reports the corresponding nutritional information. Such a system usually involves four stages: food item detection/segmentation, food type recognition, volume estimation and nutritional content assessment. Volume estimation is crucial to the whole process, since the first two stages can be performed semi-automatically or even manually by the user, while the last stage is often a simple database lookup. In this paper, we propose a complete and robust method to solve the difficult task of food portion estimation for dietary assessment using images. The proposed method is highly automatic, uses a limited number of parameters, and achieves high accuracy for a wide variety of foods in reasonable time. This study also provides an extensive experimental investigation of the choices made and a comparative assessment against state-of-the-art food volume estimation methods, on real food data.

The remainder of this paper is structured as follows: Section II outlines the previous work in 3D reconstruction and food volume estimation on mobile devices; Section III presents the proposed volume estimation method; Section IV describes the experimental setup; Section V discusses experimental results; and our conclusions are given in Section VI.

## II. Previous Work

There have been several recent attempts to automatically estimate food volume using a smartphone. To achieve this, the proposed systems have to first reconstruct the food's three-dimensional (3D) shape by using one or more images (views) and a reference object to obtain the true dimensions. Although 3D reconstruction has been intensively investigated in recent years, adapting it to such specific problems is not trivial. In the next sections, we provide an outline of basic 3D reconstruction approaches, followed by a description of the related studies on food volume estimation.

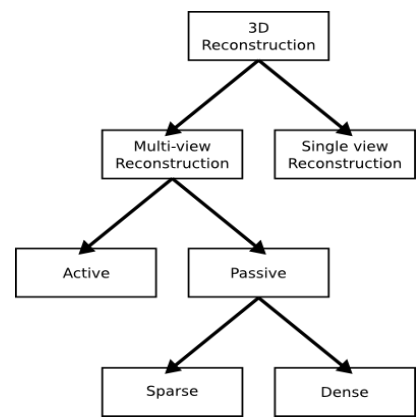

Fig. 1. Categorization of 3D reconstruction methods.

## A. 3D Reconstruction from Images

Fig. 1 presents a broad categorization of existing methods for 3D reconstruction from images. We first distinguish between methods that use a single view and those that use multiple views as input. A single view on its own does not carry sufficient information about the 3D shape, so that strong assumptions have to be made for the reconstruction. Multiview approaches however can perceive depth directly using active or passive approaches. Active approaches emit a signal and process how the scene reflects it, while passive approaches rely on changes in the signal emitted by the scene itself. Because active methods require special hardware, the present paper will focus on the description of passive systems, which require only simple cameras, such as those available on mobile devices.

The first step in all 3D reconstruction methods is extrinsic calibration, which includes the extraction of the geometric relations between the different views (relative pose) or between the views and a reference object (absolute pose). Extrinsic calibration is usually made by detecting and matching common visual structures between images or objects, followed by a model fitting paradigm. Common structures typically refer to image corners or blobs, namely salient points such as Harris corners [12], the Scale Invariant Feature Transform (SIFT [13]), or faster alternatives [14]-[15]. Out of the two sets of detected salient points, a few pairs are selected according to their visual distance and considered as common points or matches. For model fitting, the most popular choice is the RANdom Sampling And Consensus (RANSAC) family of methods [16]. RANSAC-based methods randomly sample the matches to generate candidate models and select the model with the maximum number of consenting matches, which also defines the number of remaining models to test [17].

In single-view reconstruction, the absolute pose of the camera is extracted and the 3D shape obtained through shape priors i.e. given shapes adapted to fit in the scene. In multi-view approaches the extrinsic calibration gives rise to a low resolution, sparse reconstruction of the scene (Fig. 1), where each point match between two images generates a 3D point. Dense reconstruction in multi-view methods (Fig. 1) goes further and uses all available pixels to build a 3D model. Several paths exist to dense reconstruction from images, such as shape-from-silhouette, which intersects volumes obtained by unprojecting the object's silhouette. Stereo matching methods are more common however. Dense stereo matching takes advantage of the constraints defined by the relative camera poses to simplify one-to-one pixel matching between images. These constraints serve to transform the image pairs to make point correspondences lie on the same row, a process called rectification. A literature review on stereo matching can be found in [18] together with comparative performance evaluation.

## B. Food Volume Estimation

The first systems for food volume estimation exploited the simplicity of single-view reconstruction, a trend still present nowadays. In [19], the authors propose a single-view method calibrated using a round dish of known size as reference. There, the user segments and classifies the food items, with each class having a dedicated shape model. All model parameters are refined to fit the boundaries of the food items, and the resulting parameters and model determine the volume. The required user input is burdensome however, while the instability of the results indicates that generalization is difficult.

A similar method was proposed in [20], and compared to a shape-from-silhouette algorithm [21]. The comparison, made on four dummy food items with known volume, used 20 images per item for shape-from-silhouette, and 35 images per item for single view measurements. In the latter case, all 35 estimates were averaged before evaluation, contradicting the single view principle. Both methods shower similar performance, while requiring a lot of data. A combined approach with both single- and multi-view was presented in [22], where two images are used as single view extractors of area, and thickness, with the user's thumb for reference. These necessary manipulations are difficult, and increase the chance of introducing user errors, while the volume formula remains simplistic.

The DietCam system [23] uses a sparse multi-view 3D reconstruction with three meal images. SIFT features are detected and matched between the images to calibrate and create a sparse point cloud, on which shape templates are fitted to produce the final food models. The lack of strong texture on foods results in low-resolution sparse point clouds, making post-processing ineffective. Puri et al. [24] proposed a dense, multi-view method to estimate volume using a video sequence and two large reference patterns. In this sequence, the authors track Harris corners [8], extract three images, and produce point matches between them from the tracks. The relative poses are then estimated for each image pair using a RANSAC derivative, and a combination of stereo matching and filtering generates a 3D point cloud. Finally, the point cloud is used to generate the food surface, and obtain the volume. Although this method exploits the benefits of dense reconstruction, it requires careful video capture, and a reference card the size of the dish, making it difficult to use in practice.

Despite the several attempts made recently to estimate food volume on mobile devices, they all rely on strong assumptions or require large user input. Furthermore, the published works provide limited information on algorithmic choices and tuning. The present study makes the following contributions to the relatively new field of computer vision-based dietary assessment:

- An accurate and efficient method for food volume estimation based on dense two-view reconstruction; the method is highly automatic, data-driven, and makes minimal assumptions on the food shape or type.
- An extensive investigation of the effect of parameters and algorithmic choices.

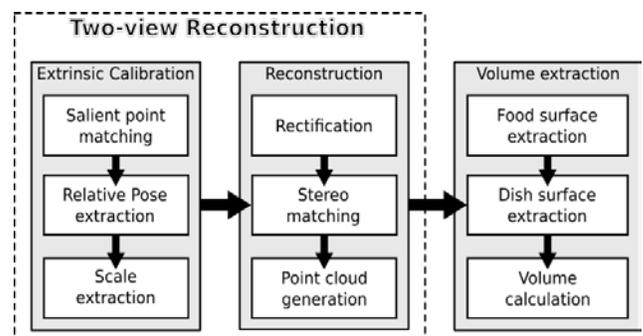
Fig. 2. Flowchart of the proposed system.

- A comparative assessment of the 3D reconstruction state-of-the-art tools for food volume estimation on real food data;

## III. METHODS

In the present work, we estimate the volume of multi-food meals with unconstrained 3D shape using stereovision (Fig. 2). The method requires two meal images with the food placed inside an elliptical plate, a credit card sized reference card next to the dish, and a segmentation of the food and dish available, possibly performed by automatic methods [25]. The dish may have any elliptical shape and its bottom should be flat. The proposed system consists of three major stages: (i) extrinsic calibration, (ii) dense reconstruction, and (iii) volume extraction.

The proposed system, like every computer vision system, relies on certain input assumptions that define its operating range and limitations. Thus, to ensure a reasonable accuracy, input images must display:
- Good camera exposure and focus.
- Strong, widespread, variable texture in the scene and limited specular reflections.
- Specific motion range between the images relative to the scene size.

The first constraint guarantees an informative scene representation and is very common in computer vision. The second constraint ensures the detection of sufficient salient points for extrinsic calibration and enough signal variation for dense reconstruction. In practice, the vast majority of foods provide enough texture, and the strong texture of the reference card also serves to fulfill this assumption. For some textureless food types like yoghurt however, the method may fail like any other passive reconstruction method. Specular reflections generate highlights that move independently of food texture, and thus falsify dense reconstruction. The third condition compromises between the ease of matching image regions in small relative motions, and the numerical precision obtained by matching locations in large relative motions. This last constraint affects both extrinsic calibration and dense reconstruction, but it can be easily resolved by guiding the user to find the optimal angles based on the phone's motion sensors.

### A. Extrinsic Calibration

Since most smartphones are equipped with a single forward facing camera, the system assumes the two images do not have fixed relative positions, and the two views must be extrinsically calibrated. The proposed calibration is performed in three steps: (i) salient point matching, (ii) relative pose extraction and (iii) scale extraction (Fig. 3).

*1) Salient point matching*

Finding point matches between the images requires first the detection and description of salient points. The first defines which parts of an image are salient, while the second describes the salient parts in a common format for comparison. For this purpose, SIFT [13], SURF [14], and ORB were considered [15]. As with many other methods, the principal tradeoff is between efficiency and accuracy, and SURF was chosen for its good combination of the two after evaluation. Once the salient points have been detected and described in both images, pairs of points are matched between the two images by comparing the descriptors in a symmetric k-nearest-neighbor search paradigm, powered by hierarchical search on k-d trees [26]. For each detected point in the two images, the top ranking match in the other image is obtained and the set of directional matches is intersected to find symmetric cases (Fig. 3(a)).

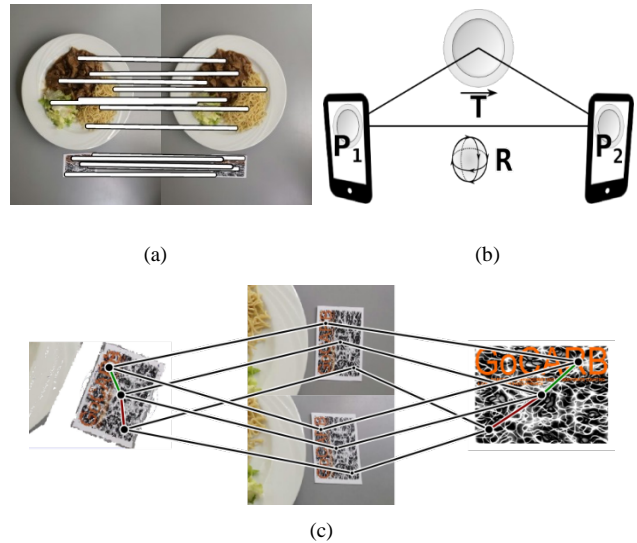

Fig. 3. Extrinsic calibration: (a) salient point matching, (b) pose extraction, (c) scale extraction (the ratio of lengths between segments of the same color creates scale candidates).

*2) Relative pose extraction*

To extract the relative pose model (Fig. 3(b)), a RANSAC-based method serves as a basis [16], followed by iterative refinement of the solution. At each iteration of the algorithm, five matches are sampled at random. From this sample set, models are created using the five point relative pose model generator of Nister [27]. The generated models are evaluated on all matches and scored according to the proportion of inliers, i.e. the inlier rate. Inliers are matches for which the symmetric epipolar distance [28] to the model is below a given threshold - the inlier threshold. The model with the largest inlier rate is saved until a better one is found. The inlier rate of the best model redefines the maximum number of iterations; after which the algorithm terminates and we remove outliers from the match set. The resulting model is then iteratively refined using the Levenberg Marquardt (LM) optimization algorithm on the sum of distances [29]. Finally, a sparse point cloud is created from the inlier matches through unprojection.

The classical RANSAC algorithm was modified by including i) local optimization, and ii) a novel, adaptive threshold estimation method. The implemented approach is presented in Fig. 4, and the modifications made in the classical RANSAC are described in the following paragraphs.

*Local Optimization:* When a new best model is found, the RANSAC algorithm goes through local optimization (LO) [17]: sets of ten matches are randomly sampled from the inliers of the best model to generate new, potentially better models. LO uses the fitness function of (1) (from MLESAC [18]) instead of the inlier rate, to score the models, which explicitly maximizes inlier count and minimizes inlier distances. To further improve on this principle, when a new best model is found in LO, we restart the local optimization using this model's inlier set, thus testing more high-quality models.

$$\text{Inlier\_fitness}(M,D) = \sum_{p \in D} \text{Max}(\text{Thr}_{inl} - \text{dist}(M,p), 0) \quad (1)$$

where M is a model, D is the set of matches, $\text{Thr}_{inl}$ is the inlier threshold and dist( ) is the symmetric epipolar distance.

*Adaptive threshold estimation:* Instead of using a fixed threshold between inliers and outliers, we determine it for each image pair and each best model, through an a-contrario method influenced by [30]. In our system, we create a noise distribution from the data by randomly matching existing points together, and we use this distribution to find the appropriate threshold and filter out outliers. In the main loop, a threshold is estimated locally for each model until a non-trivial model is found. Once such a pose model is found, the same method is used to define a global threshold, which is updated with each better model.

To find the threshold for a given model, let $\text{Cdf}_{data}: [0; \infty[ \to [0; 1]$ be the cumulative distribution function of distances to the input matches, and $\text{Cdf}_{noise}: [0; \infty[ \to [0; 1]$ the cumulative distribution function of distances to the random matches. For a given inlier threshold T, the ratio $\text{FDRB}(T) = \text{Cdf}_{noise}(T)/\text{Cdf}_{data}(T)$ is the maximum percentage of inliers to the model that can be produced by random matches: it is an upper bound of the false discovery rate. This formulation is equivalent to $\text{FDRB}'(\delta) = \text{Cdf}_{noise}(\text{Cdf}_{data}^{-1}(\delta))/\delta$, where $\delta = \text{Cdf}_{data}(T)$ is the inlier rate. This parametrization abstracts the distance function, and makes the input and output of the function meaningful rates of the data. The chosen threshold gives the largest inlier rate while keeping the false discovery rate bound below a fixed value p: $T_{opt} = \text{Cdf}_{data}^{-1}(\text{argmax}\{\delta, \delta \in [0; 1[, \text{FDRB}'(\delta) < p\})$. Here, p, the maximum noise pollution rate, is set at 3% after experiments.

*3) Scale estimation*

Until now, the relative pose between two cameras was defined up to an arbitrary scale factor. The true scale is extracted from the matched points on the reference card (Fig. 3(c)). Each match consists of one salient point on each image and both points are matched to the reference card to build triangular point matches (tracks). Besides, the two image points in each track correspond to a 4th point in the sparse cloud. Because the reference card is flat, the matches coincide with a 2D projective transform - a homography - between the images and the original reference card pattern. The homography between the first image and the reference card is extracted using RANSAC with an inlier distance threshold 3% of the reference card length. Finally, the ratios between each distance in the sparse point cloud and its equivalent in the reference pattern are estimated, and the mode of their distribution is chosen as the scale.

### B. Dense Reconstruction

For the dense reconstruction, we use a stereo matching approach involving three steps: (i) rectification of the images (Fig. 5(a)), (ii) stereo matching (Fig. 5(b)) and (iii) point cloud generation (Fig. 5(c)).

*1) Rectification*

Rectification simplifies dense matching of points between the two images by restricting the matching process between rows (Fig.5 (a)). The rectification method used in the proposed system works regardless of camera configuration [31]. This method describes the two images in a consistent radial coordinate system, where the center for each view is the projection of the other camera center, and angles are matched around it. To simplify the stereo matching, we force the salient point matches to also satisfy a consistent pairwise horizontal order. A pair of matches is marked as "consistent" if their horizontal order is the same in the rectified images, and "inverted" otherwise. If more than half the matched pairs are marked as "inverted", the second rectified image is mirrored horizontally.

*2) Stereo Matching*

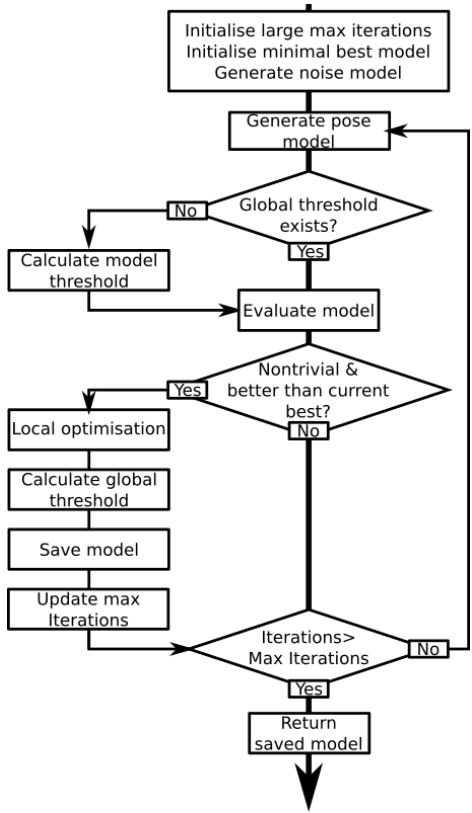

Fig. 4. RANSAC outline, including adaptive threshold estimation and local optimization.

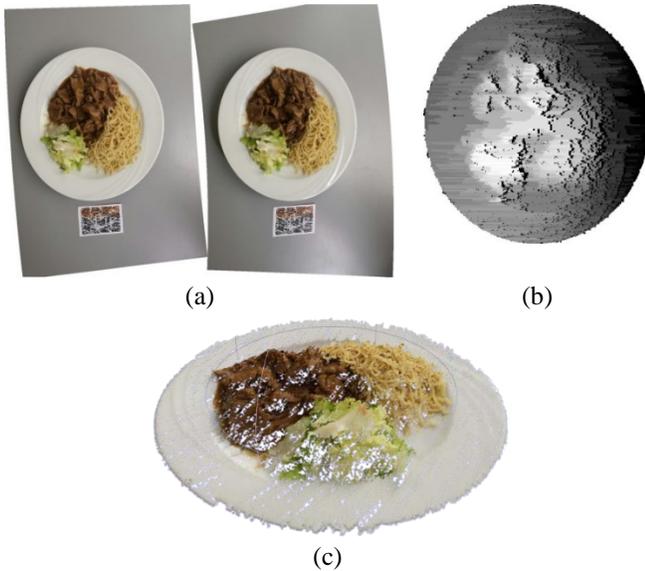

Fig. 5. Dense reconstruction with stereo matching (a) rectification, (b) intensity coded horizontal motion of pixel matches (disparity map), (c) dense unprojected point cloud.

Stereo matching is the process of finding dense pixel matches along corresponding rows of rectified images by comparing image patches around each pixel, which results in a disparity map (Fig.5(b)). Matching can be described by four parts [18]: matching cost, aggregation, optimization, and refinement. To these, we add the extraction of the disparity range.

*Disparity range*: The expected range of horizontal motion between points in the two rectified images. Fig. 5(b) shows the variation in disparity for a given image pair. The disparity range is extracted dynamically from sparse point matches: after rectifying the matches the central 90% of the disparities for these matches creates a robust tentative range. The range is enlarged by a fixed amount to ensure it contains the true search range.

*Matching cost:* A dissimilarity measure between image patches. Here, we use the Census filter [32], which proved the best choice after comparative experimentation. The Census filter describes local texture using Local Binary Patterns (LBP), which we compare with each other using the hamming distance. In LBPs, each value of a rectangular area is compared to the value at the center of the rectangle and binarized. A greater value is assigned a 1 in the descriptor, and a lesser value 0. Because the Census filter keeps only binary differential information, it is robust to monotonous transformations of the values, and for that same reason it gives the highest performance [33]. Thus, given the assumption of a sufficiently textured scene, the proposed dense matching approach performs satisfactorily.

*Aggregation*: The shape and size of the image patches that are compared between the two images. In this implementation, a square window of fixed size is used.

*Optimization*: The integration of local matching cost into a global framework for piecewise continuous matching. For this, we use a hierarchical dynamic programming approach based on [34]. Dynamic programming solves the stereo matching problem by calculating a set of row to row matchings. These matchings are calculated by formulating the problem as path-optimization within a matrix. Each entry in the matrix corresponds to the cost of matching a pixel in the first image to one in the second image. Dynamic programming seeks the path with the lowest length-regularized cost. Here, the regularization cost function is inversely related to the local variance, to enforce smoothness in visually smooth areas. Hierarchical stereo matching effectively improves performances both in speed, by using the largest search spaces on the smallest image pairs, and in accuracy, by searching for a solution at multiple scales. For additional robustness against scale differences, the two images are first resized to a fixed dish size.

*Refinement*: Post-processing steps to increase matching accuracy. After computing the disparity map, we apply a median filter to smooth the result.

3) Point cloud generation

The 3D positions of dense matches are extracted by transferring stereo matches back to the original image coordinates using the inverse transformation of the rectification. Each match is then unprojected directly using the relative pose to generate 3D points (Fig. 5(c)), and the depths are integrated into a depth map for the first image.

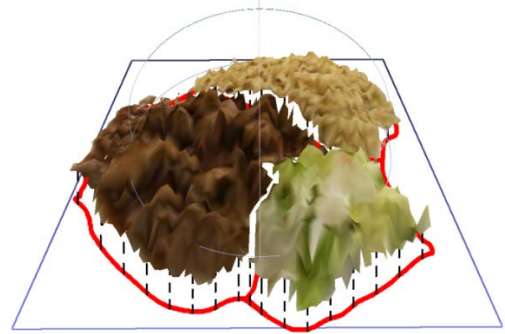

Fig. 6. Volume estimation: food surface (separated using the segmentation map), plate surface and vertical projection on the dish plane.

C. Volume Estimation

The third major stage of the proposed system is to estimate the volume of the food based on the reconstructed point cloud. Before calculating the volume, the surface of the food needs to be "closed" into a watertight shape and for this we generate the dish surface lining its bottom. Fig. 6 shows the food surfaces, and their projection on the dish surface during integration.

1) Food Surface Extraction

Using the available segmentation map, the background is eliminated, and the shape is sampled on a regular grid to generate a mesh with a fixed total number of points. Then, using the 2D image coordinates of the sampled points, the surface is partitioned into triangles by Delaunay triangulation [35]. Triangles belonging to different segments are discarded, producing a set of distinct surfaces.

2) Dish Surface Extraction

Most dishes of plastic or glazed ceramic are smooth and reflective. These properties make dishes difficult to reconstruct using passive reconstruction, if at all possible. Because of this, it is currently safer to rely on common plate properties than to attempt online fitting. Standard practice consists in using a plane shifted above the table to represent the "bottom" of the dish [19], [24]. The plane equation is then defined using the reference card, and the height of the plate surface. The accuracy of this method is limited by the size of the card and its location relative to the food. Here, we build the dish surface through a three-fold process combining the dish border and the reference card. First, the plane of the dish rim is extracted, second, its distance from the table is obtained from the reference card, and third, the plane is shifted to a given height above the table. To extract the 3D rim of the dish, we densely match the dish borders points, unproject those matches, and feed them to RANSAC to find the most representative plane. After this, the modal dish height is obtained using the reference card to define the table plane, and the rim plane is shifted to a fixed height above table level.

*3) Volume Calculation*

Extracting the volume is the last step in the process, and it combines the results of the previous sections. The dish surface defines the vertical direction through its normal, and so we calculate the volume by integrating the height of the food surface above the dish. This integration is simply the sum for all surface triangles of their horizontal area multiplied by the average height of their corners.

## IV. DATA AND EXPERIMENTAL SETUP

Here we describe the data used for the experiments along with the evaluation protocol and provide some details on the implementation of the proposed system.

### A. Dataset

To test the performance of volume estimation, we use four datasets. The first consists in 45 dishes (*Meals-45*) from the restaurant facilities of the Bern University Hospital, "Inselspital". Out of the 45 dishes, 23 contain three distinct food items with the rest containing two, resulting in a total of 113 items. Part of the *Meals-45* dataset contains images captured from specific angles using a rotating arm mounted on a table. This secondary dataset (*Angles-13*) contains 13 dishes and is used to show the effect of the relative shooting angle on the results. The third dataset (*Plates-18*) includes 6 meals from an international fast food retailer, served in 3 different types of plates (dinner, soup, oval). This sums to 18 dishes which are used to quantify the effect of changing plates on the result, as well as to test the method on different food types. Finally, the last dataset consists in 14 additional meals from local supermarkets and restaurants, and aims to stress the system even further and test its versatility and generalization capabilities (*Meals-14*). Fig. 7 shows example images from the different datasets.

For each dish, three image pairs were taken from distances of 40 to 60 cm, within 45 degrees from the vertical, using a Samsung Galaxy S2 Plus smartphone *(Meals-45, Meals-14)* and a Samsung Galaxy S4 smartphone *(Plates-18)*. Concurrently, a high accuracy point cloud was created and registered to the images using the ASUS XtionPRO depth sensor and the Kinect Fusion algorithm [36]. For each image pair, the dish and the different food items were manually segmented. The dominant horizontal plane (table plane) was extracted with RANSAC, shifted to the true dish height, after which the ground truth volume was extracted. The entire dataset is, to the best of our knowledge, both the largest and most complete one existing for food volume estimation.

### B. Evaluation Protocol for Food Volume Estimation

For each dish of all datasets, and all three image pairs per dish, the complete system was run eight times, giving 24 estimates for each food item. Because the ultimate goal was to calculate quantities, we used the mean absolute percentage error in volume (MAPE) to evaluate the system's accuracy over each dataset. For a food item i, $MAPE_i$ is calculated over the 24 estimates (2) while the coefficient of variation $CV_i$ (3) is used to assess the precision/stability of the method. The $MAPE_{overall}$ is calculated over all food items (4), and serves for general evaluation. Speed will also be given since it is often crucial for the final implementation.

$$MAPE_i = \frac{1}{24}\sum_{j=1}^{24}|\frac{V_i - \hat{V}_i^j}{V_i}| \qquad (2)$$

where $V_i$ is the true volume of item i, $\hat{V}_i^j$ is volume estimate j of item i

$$CV_i = \frac{1}{\overline{V}_i}\sqrt{\frac{1}{24}\sum_{j=1}^{24}(\hat{V}_i^j - \overline{V}_i)^2} \qquad (3)$$

with $\overline{V}_i = \frac{1}{24}\sum_{j=1}^{24}\hat{V}_i^j$

$$MAPE_{overall} = \sum_{i=1}^{items} MAPE_i / items \qquad (4)$$

### C. Implementation

All proposed methods were implemented in C++ without parallelism, and executed under Linux using an Intel i7-3770K (8MB Cache, 3.5GHz) CPU and 8GB of RAM. The salient point detection and matching were made using the OpenCV library version 2.4.10 [26] and openSURF [37].

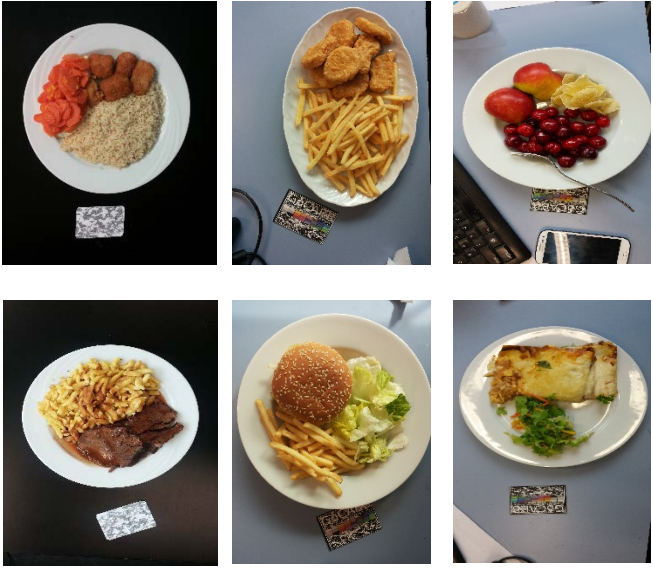

Fig. 7. Example dishes from the different datasets. The two images on the left belong to the *Meals-45* dataset, the ones in the middle come from the *Plates-18* and the two on the right are from the *Meals-14* dataset.

TABLE II
PERFORMANCE FOR THREE MAJOR SALIENT POINT DETECTOR/DESCRIPTORS

| Detector/Descriptor | $MAPE_{overall}$(%) | Average time(s) |
|---|---|---|
| ORB | 14.8 | 5.9 |
| SIFT | 9.6 | 11.6 |
| SURF | 8.2 | 5.5 |

TABLE I
PERFORMANCE OF DIFFERENT MATCHING TECHNIQUES USING SURF

| Top-ranking matches | Distance cutoff (%) | Distinctiveness ratio | $MAPE_{overall}$ (%) | Average time (s) |
|---|---|---|---|---|
| 2 | 0 | 1 | 9.4 | 9.7 |
| 1 | 0 | 1.05 | 9.2 | 7.2 |
| 2 | 50 | 1 | 9.1 | 9.3 |
| 1 | 25 | 1 | 9.1 | 6.9 |
| 1 | 0 | 1.15 | 8.75 | 6.9 |
| 2 | 0 | 1.1 | 8.5 | 6.9 |
| 1 | 0 | 1 | 8.5 | 6.4 |
| 1 | 50 | 1 | 8.4 | 7 |
| **1** | **0** | **1.1** | **8.2** | **5.5** |

## V. RESULTS AND DISCUSSION

In this section, we first present the results that justify our choices in algorithms and parameters based on the system's performance on the *Meals-45* dataset. One subsection is provided for each of the three main stages: extrinsic calibration, reconstruction, and volume extraction; benchmarking the available choices. Each experiment refers to a single modification at a time, while the rest of the system was partially optimized. Small deviations in the average error may occur due to the random nature of RANSAC, so changes below 0.2% are negligible. After the step-by-step review, the final system is presented before giving a review of its performance. The effects of the shooting angle and the used plate are investigated using the *Angles-13* and *Plates-18* datasets, respectively. Finally, the proposed method is tested on the *Meals-14* dataset to assess its generalization capabilities and compared as a whole to other available methods for food volume estimation.

### A. Extrinsic Calibration

We begin with a comparison of different methods for detecting, describing and matching salient points which is followed by the evaluation of the different choices for the relative pose and scale estimation.

*1) Salient Point Detection/Description*

Table I compares the results using ORB, SIFT, and SURF. ORB reaches the highest speed, but it also results in the highest error. SIFT is much slower, but using it reduces the error by 5.2%. Finally, SURF proved to be the best choice: it is considerably faster than SIFT, and reduces the error by a further 1.4%.

*2) Salient Point Matching*

After experimenting with salient point detectors/descriptors, we evaluated different techniques to match those points symmetrically between the two images. Symmetrical matches can be filtered by thresholding the number of top-ranking matches, the distance, or the ratio of distances between matches sharing a point i.e. distinctiveness. As can be seen from Table II, increasing the number of best ranking matches not only increases execution time, but also errors, a result of decreasing match quality. Errors also increase when eliminating a fixed percentage of the worst matches. This shows that matching distance alone may be a bad predictor of data quality compared to the ordering of distances. Enforcing distinctiveness between consecutive matches improved the accuracy of the system by 0.3%, suggesting that better separation of matches preserves the higher quality points and matches. Yet this effect is limited, and requiring large distinctiveness filters out many valid matches.

*3) Relative Pose Extraction*

Table III presents the performance of the system using different relative pose extraction techniques. Time spent in the relative pose is relatively small, with approximately 5% of total execution time therefore we focus on gains in accuracy. Our version of Local optimization (LO) reduced the error by 2.5% compared to standard RANSAC, and the LM minimization a further 1.2%. Using an adaptive inlier threshold instead of static improved the results further, indicating that the appropriate inlier threshold may vary widely in the dataset. The threshold estimation method of [30] reduced the error by 1.1% while the proposed dynamic threshold with the same setup resulted in an additional .4% decrease in the error, making this last combination the most accurate of all those examined here for no additional cost. Furthermore, we tested the PRESAC variant of RANSAC as used in [24] with a much larger number of iterations. It proved inferior to the proposed method and more difficult to tune since more a priori knowledge of data quality is required to define the number of iterations.

Finally, for the detection of the reference card, the error of the system was negligibly lower when using calibrated absolute pose models (8.1%) than when using homographies (8.2%), indicating that the problem is easily solved without additional constrains.

TABLE III
PERFORMANCE OF THE SYSTEM FOR THE DIFFERENT RELATIVE POSE EXTRACTION TECHNIQUES

| Relative Pose Extraction Method | | | | $MAPE_{overall}$ (%) |
|---|---|---|---|---|
| Robust fitter | LO | Threshold | LM | |
| RANSAC | No | Static | No | 13.4 |
| RANSAC | Yes | Static | No | 10.9 |
| RANSAC | Yes | Static | Yes | 9.7 |
| RANSAC | Yes | Adaptive [30] | Yes | 8.6 |
| **RANSAC** | **Yes** | **Proposed** | **Yes** | **8.2** |
| PRESAC [24] | No | Static | Yes | 9.1 |

### B. Dense Reconstruction

The following components and parameters were evaluated for reconstruction: the matching resolution, the delimitation of the region of interest, and the matching cost. The matching resolution is the size at which stereo matching is applied, and may be different from the original image resolution. Experiments showed that the method is fairly insensitive to matching resolution above a reasonable minimum value. Therefore, we rescaled the image to a fixed dish area of $2^{17}$ pixels to achieve the most efficient solution. Excluding the background from the matching region of interest not only increased the speed by 22%, but it also reduced the error (8.2% over 9.6%) Finally, as matching cost we chose the Census filter that performed best by far with a negligible increase in execution time. Using alternative costs like the sum of absolute differences (SAD) or the zero-mean normalized cross correlation (ZNCC) resulted in an increase of 4.5% and 5.5% in the error, respectively.

TABLE IV
COMPARISON OF DIFFERENT 3D MESH SIZES

| Mesh size (total number of points) | MAPE$_{overall}$ (%) | Average time(s) |
|---|---|---|
| $2^{10}$ | 9.9 | 5 |
| $2^{11}$ | 8.5 | 5.2 |
| $2^{12}$ | 8.2 | 5.5 |
| $2^{13}$ | 8.8 | 6 |
| $2^{14}$ | 9.3 | 12 |

TABLE V
COMPARISON OF SYSTEM RESULTS USING DIFFERENT RELATIVE ANGLES WITH A FIXED FIRST IMAGE (*ANGLES-13* DATASET)

| Fixed angle (degrees) | MAPE$_{overall}$ (%) | Average time(s) |
|---|---|---|
| 5 | 8.4 | 4.2 |
| 10 | 6.0 | 4 |
| 15 | 5.1 | 3.9 |
| 20 | 5.1 | 3.9 |
| 25 | 6.3 | 4 |
| 30 | 6.3 | 4.1 |
| 35 | 30.4 | 4.3 |

*C. Volume Extraction*

Table IV shows the effect of using different resolutions for the reconstructed food surface. The error decreases as the resolution of the mesh increases up to $2^{12}$. The comparatively low performance of higher resolutions appears to show that the mesh captures more noise, held in the high frequencies of dense matching. Since increasing resolution also increases processing time, we use point cloud sizes of $2^{12}$ points. To define the dish surface, the proposed combined use of the plate's rim and the reference card proved significantly more accurate than using the reference card alone, with little to no effect on execution speed. Using the latter resulted in an average error of 15%, nearly 7% higher than the proposed.

*D. Evaluation of the Proposed System*

After evaluating and tuning the involved components, the proposed food volume system can be summarized. For the detection/description of salient points we use SURF, which are then matched using single top match intersection with a distinctiveness ratio of 1.1. The relative pose is extracted using the proposed RANSAC method, LO on the inlier fitness criterion, dynamic thresholding based on the FDRB and finally iterative LM optimization. Dense reconstruction is made on the content of the dish using hierarchical dynamic programming, where the smoothness cost is a function of local variance. Finally, the dish surface is obtained by combining the dish border and reference card information. The proposed system achieved an MAPE$_{overall}$ equal to 8.2% on the *Meals-45* dataset. In this section, we give additional insight into the system's performance, first by looking at the distributions of errors and the coefficient of variation of estimates, then at the effect of the relative angle between the two views, and finally by comparing it to other similar methods of volume estimation.

*1) Analysis of Errors*

So far, we have used the MAPE$_{overall}$ to guide evaluation. Here, we derive finer grained information, first by drawing the distribution of the signed percentage errors on the *Meals-45* dataset for all 113 food items and 24 estimates (= 2712 values). The distribution (Fig. 8(a)) looks fairly smooth and symmetrical with its peak and mean around 0. Sixty-nine percent of the samples have a signed percentage error within ±10%, and 91.5% lie within ±20%. Some outliers with large errors can also be observed, as caused by low input quality. Fig. 8(b) presents the distribution of CVi (3) for all the 113 different food items of *Meals-45*. The distribution of CVi has an average value of 7.1% with 81.8% of values under 10%, and 99.2% under 20%. This distribution shows that while most items are stable, a few display high variability, mainly caused by the same low quality input as before.

*2) Optimal Angle Estimation*

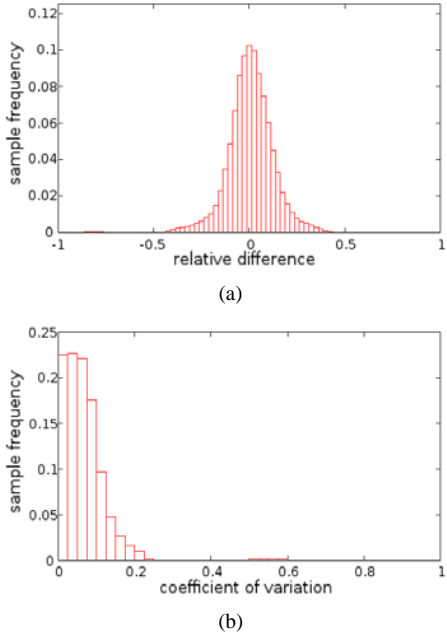

Fig. 8. (a) Distribution of signed percentage error, (b) distribution of the coefficient of variation of estimates per item (3).

TABLE VII
COMPARISON OF RESULTS FOR DIFFERENT PLATES

| Plate | $MAPE_{overall}$ (%) |
|---|---|
| Dinner | 8.9 |
| Soup | 6.4 |
| Oval | 7.1 |

TABLE VI
COMPARISON OF EXISTING FOOD VOLUME ESTIMATION PRINCIPLES

| Method | $MAPE_{overall}$(%) on Meals-45 | $MAPE_{overall}$(%) on Meals-14 | $MAPE_{overall}$(%) on Plates-18 | Average time(s) |
|---|---|---|---|---|
| Single view | 23.1 | 30.8 | 29.8 | 3.5 |
| Sparse Multi-view | 21.1 | 18.5 | 13.9 | 4 |
| Dense Multi-view | 14.9 | 15.5 | 11.5 | 5 |
| Proposed | 8.2 | 9.8 | 7.4 | 5.5 |

Table V shows the $MAPE_{overall}$ for food items of the dataset *Angles-13* with fixed relative angles and a fixed distance from the dish. At relative angles of 15 and 20 degrees, the results reach a double minimum: in average error and average execution time. The constant increase in error and time outside of 15-25 degrees indicates that a relative angle in this range results in stable executions and should be chosen when capturing meals. This experiment confirms known tradeoffs of relative angle and motion on reconstruction: smaller angles and motion make matching easier, but the accuracy of depth estimation is low, and conversely. This also indicates that better results than those presented for *Meals-45* could be obtained by enforcing specific properties on the camera configuration, either through hardware, or user guiding systems.

*3) Plate effect*

The *Plates-18* dataset allows us to evaluate the effect of the dish shape on the results as well as test the proposed system on different food types. The average performance of the system on three different dishes is shown in Table VI. Two things are apparent: (i) the average error is close to that of the *Meals-45* dataset, and (ii) the variation in the average error is less than 2.5% across plates. The first point illustrates that the software extends from the original dataset to this one despite the differences. The second point shows instead that the results are robust to the plate style.

*4) Testing on a different dataset*

Here, we test the method on the *Meals-14* dataset, which contains foods from different sources compared to all previous datasets. These are fast-food dishes, supermarket ready-to-eat food, and fruits. Using this extra set of data we can evaluate how well the system generalizes to foods from other sources. The $MAPE_{overall}$ on this new dataset was 9.8%. Comparing this result with the 8.2% over the *Meals-45* dataset, we can see a small increase in error, but overall the results remain stable. This shows that although the system does not perfectly fit the variety of foods, it can approximate their volumes well enough without being tuned on them.

*5) Comparison to other methods*

There are few methods to estimate food volumes from images in the literature. Those that exist cluster around three groups: single view with priors, of which we take [20] as representative, sparse multiple-view such as [23], and dense multi view like [24]. Because the latter uses more inputs we show results for a two-view adaptation of the work. Where implementation information was unavailable, we used the highest performing method available within our method. Table VII presents the comparison of the proposed with the aforementioned methods on the different datasets.

The first two approaches achieved an average error more than twice that of the proposed method on the *Meals-45* dataset, while the other dense reconstruction system followed with an error nearly twice higher. On the other two datasets, single view reconstruction was even more inaccurate, which was expected because of the gross shape approximations that reduce the method's generalization ability. The alternative dense method performed better than the rest but it was still inferior to the proposed. Regarding processing times, although both single view and sparse 3D are generally faster, their low accuracy makes their use in such an application prohibitive.

## VI. CONCLUSION

In this paper, we presented an algorithmic framework for 3D reconstruction and volume estimation of various food items contained on a dish. In this complex problem, parameters influencing the extrinsic calibration had the greatest effect on the results. The system outperformed existing solutions having overall mean absolute percentage error ranging from 8.2% to 9.8% on the different datasets, with over 90% of volume estimate errors under 20%, all in roughly 5.5 seconds. The proposed method is currently part of a carbohydrate counting system for individuals with diabetes [38], and may also be used for general food assessment. It uses segmentation to calculate portion sizes, and, by using food recognition, the nutrient profile is built up. The system has been tested in a pre-clinical trial showing greater accuracy than the target users [39], and in a clinical trial which showed a positive impact to the self-management of type 1 diabetes [40].

Moreover, the present system may benefit from a number of improvements. The relative pose extraction can benefit from more advanced distances between data and model, as well as more distinctive noise models. Dense reconstruction can be improved with more complex matching algorithms, at the cost of speed. As a final improvement, we may expand the system to more complex food containers models for dishes like bowls, or reconstruct the shape of the food container from the images.

## REFERENCES

[1] World Health Organization (2014). Obesity: Situation and trends. [Online]. http://www.who.int/gho/ncd/risk_factors/obesity_text/en/
[2] International Diabetes Federation (IDF) [Online] - http://www.idf.org/
[3] Livingstone MB et al., "Issues in dietary intake assessment of children and adolescents," in British J. Nutrition, vol. 92, pp213–S222, Oct 2004.
[4] Godwin SL et al. "Accuracy of reporting dietary intake using various portion-size aids in-person and via telephone," in J. American Diet Assoc., vol.104, n.4, pp. 585–594, Apr 2004.

[5] Beasley J et al., "Accuracy of a PDA-based dietary assessment program," in Nutrition, vol. 21, no. 6, pp. 672-677, Jun 2005.
[6] Hernández T et al., "Portion size estimation and expectation of accuracy," in J. Food Composition and Anal., vol. 19, pp. 14-21. 2006.
[7] Schoeller, D., et al., "Inaccuracies in self-reported intake identified by comparison with the doubly labelled water method," in Canadian J Physio, Pharmaco,, vol. 68, no. 7, pp.941-945, Jul 1990.
[8] M. Graff, T. Gross, S. Juth, and J. Charlson, "How well are individuals on intensive insulin therapy counting carbohydrates?" in Diabetes Research and Clinical Practice, vol. 50, pp. 238-239, Sep 2000.
[9] M.C. Carter, V.J. Burley, C. Nykjaer, and J.E. Cade. "My Meal Mate: validation of the diet measures captured on a smartphone application to facilitate weight loss," in British J. Nutr, vol. 3, pp. 1-8, 2012.
[10] S. Kikunaga, T. Tin, G. Ishibashi, D.H. Wang, S. Kira. "The application of a handheld personal digital assistant with camera and mobile phone card (Wellnavi) to the general population in a dietary survey," in J. Nutrition Sci. Vitaminol, vol. 53, pp. 109-116, 2007.
[11] C.K. Martin, S. Kaya, B.K. Gunturk, "Quantification of food intake using food image analysis,§ in Conf. Proc. IEEE Eng. Med. Biol. Soc., Minneapolis, MN, USA, Sep. 2009, pp. 6869–6872.
[12] C. Harris, M. Stephens, "A combined corner and edge detector," in Proc. Of the 4th Alvey Vision Conf., Manchester, UK, 1988, pp-147-151.
[13] D.G. Lowe, "Distinctive image features from scale-invariant keypoints," in Int. J. of Comput. Vision, vol. 60, no. 2, pp. 91-110, November 2004.
[14] H. Bay et al, "SURF: Speeded Up Robust Features," in Comput. Vision and Image Understanding, vol. 110, no. 3, pp. 346-359, Jun 2008.
[15] E. Rublee, et al., "ORB: an efficient alternative to SIFT or SURF," IEEE Int. Conf. on Comp. Vision, Barcelona, Nov 2011, pp. 2564-2571.
[16] Fisher, M., Bolles, R., "Random sample consensus: A paradigm for model fitting with applications to image analysis and automated cartography," in Comm. of the ACM, vol. 24, no. 6, pp. 381–395, 1981.
[17] S. Choi et al. "Performance Evaluation of RANSAC Family," in Proc. British Machine Vision Conf., London, UK, Sept. 2009, pp. 81-92.
[18] D. Scharstein and R. Szeliski, "A taxonomy and evaluation of dense two-frame stereo correspondence algorithms," in Int. J. Comput. Vision, vol. 47, pp. 7-42, Apr-Jun 2002.
[19] H-S Chen, W. Jia, Y. Yue, Z. Li, Y.-N. Sun, J.D. Fernstrom, M. Sun, "Model-based measurement of food portion size for image-based dietary assessment using 3D/2D registration," in Measurement Science and Technology, vol. 24, pp. 1-11, 2013.
[20] Xu, Y. He, N. Khanna, C. Boushey, E. Delp, "Model-based food volume estimation using 3D pose," in Proc. IEEE International Conf. on Image Processing, Melbourne, Australia, September 2013.
[21] A.W. Fitzgibbon, G.Cross, A. Zisserman, "Automatic 3D model construction for turntable sequences," in Lecture Notes in Comp. Science, vol. 1506, pp. 154-170, 1998.
[22] R. Almaghrabi, G. Villalobos, P. Pouladzadeh, S, Shirmohammadi, "A novel method for measuring nutrition intake based on food image," in. IEEE Int. Conf. on Instrumentation and Measuring Technology, Graz, 2012, pp. 366-370.
[23] F. Kong, J. Tan, "DietCam: automatic dietary assessment with mobile camera phones," in J. Pervasive and Mob. Computing, vol. 8, no. 1, pp. 147–163, Feb. 2012.
[24] M. Puri, et al., "Recognition and volume estimation of food intake using a mobile device," in Proc. IEEE Workshop on Applications of Comp. Vision, Sunbird, UT, USA, December 2009, pp. 1–8.
[25] J. Dehais, M. Anthimopoulos, and S. Mougiakakou, "Dish Detection and Segmentation for Dietary Assessment on Smartphones", ICIAP 2015 Workshops, vol. 9281, pp. 433–440, 2015.
[26] Open Computer Vision Library [Online]: http://www.opencv.org
[27] Nistér, "An efficient solution to the five-point relative pose problem.", in IEEE Transactions on Pattern Analysis and Machine Intelligence, vol. 26, no.6, pp 756-770, 2004.
[28] Hartley R. I. and Zisserman A. Multiple View Geometry in Comp. Vision III edition, (2004), "Cambridge University Press",
[29] D. Marquardt, "An algorithm for least-squares estimation of nonlinear parameters", in Journal on Applied Mathematics, 11(2), pp. 431–441.
[30] L. Moisan, P. Moulon, P. Monasse, "Automatic homographic registration of a pair of images, with A contrario elimination of outliers", in Image Processing On Line, vol 2,pp. 56–73, 2012.
[31] M. Pollefeys, R. Koch, L. Van Gool, "A simple and efficient rectification method for general motion," in IEEE Int. Conf. Comp. Vision, Corfu, Greece, September 1999, vol.1, pp. 456 – 501.
[32] R. Zabih, J. Woodfill, "Non-parametric local transforms for computing visual correspondence," in Lecture Notes in Comp. Science, 801 (1994), 151-158e
[33] H. Hirschmueller, D. Scharstein, Evaluation of stereo matching costs on images with radiometric differences, IEEE Transactions on Pattern Analysis and Machine Intelligence, 31 (9) (2009), 1582-1599
[34] G. Van Meerbergen, M. Vergauwen, M. Pollefeys, L. Van Gool. "A hierarchical symmetric stereo algorithm using dynamic programming," in Int. J. on Comp. Vision, 47 (2002), 275-285.
[35] M. De Berg, O.Cheong, M. van Kreveld, M. Overmars, "Computational geometry; algorithms and applications," Springer Verlag.
[36] R.A. Newcombe, S. Izadi, O. Hilliges, D. Molyneaux, D. Kim, A.J. Davison, P. Kohli, J. Shotton, S. Hodges, A. Fitzgibbon, "KinectFusion: real-time dense surface mapping and tracking," in: IEEE International Symposium on Mixed and Augmented Reality, Basel, Switzerland, October 2011, pp. 127-136.
[37] OpenSURF [Online]: https:code.google.com/p/opensurf1/
[38] M. Anthimopoulos, J. Dehais, S. Shevchik, R.H. Botwey, D. Duke, P. Diem, S. Mougiakakou, "Computer vision-based carbohydrate estimation for type 1 diabetic patients using smartphones," in J. Diabetes Sci Technol Apr. 2015.
[39] D. Rhyner, H. Loher, J. Dehais, M. Anthimopoulos, S. Shevchik, R.H. Botwey, D. Duke, C. Stettler, P. Diem, and S. Mougiakakou, "Carbohydrate Estimation by a Mobile Phone-Based System Versus Self-Estimations of Individuals With Type 1 Diabetes Mellitus: A Comparative Study, " in J. Medical Internet Research, vol. 18, no. 5, 2016.
[40] L. Bally, J. Dehais, C.T. Nakas, M. Anthimopoulos, M. Laimer, D. Rhyner, G. Rosenberg et al. "Carbohydrate Estimation Supported by the GoCARB system in Individuals With Type 1 Diabetes: A Randomized Prospective Pilot Study." in Diabetes Care, 2016.